%% file: main.tex
\documentclass{article}

\usepackage[nonatbib, final]{neurips_data_2021}
\usepackage[numbers]{natbib}
\usepackage[utf8]{inputenc} % allow utf-8 input
\usepackage[T1]{fontenc}    % use 8-bit T1 fonts
\usepackage{hyperref}       % hyperlinks
\usepackage{url}            % simple URL typesetting
\usepackage{amsfonts}       % blackboard math symbols
\usepackage{nicefrac}       % compact symbols for 1/2, etc.
\usepackage{microtype}      % microtypography
\usepackage{enumitem}% so we can change itemize indent
\usepackage{todonotes}
\usepackage{listings}  % For nice algorithm code
\usepackage{bm}
\usepackage{xltabular}
\usepackage{pifont}
\input{_packages_macros/table_pckgs}

\usepackage{graphicx}
\graphicspath{{./figures/}}
\usepackage{wrapfig}
\usepackage[font=small,skip=0pt]{subcaption} %subfigure
\usepackage{balance}

\usepackage{amsmath,soul,amssymb}

\title{ClimART: A Benchmark Dataset for Emulating Atmospheric Radiative Transfer in \\ Weather and Climate Models}

\author{%
  Salva Rühling Cachay\thanks{Equal contribution} \\
   TU Darmstadt \& Mila\\
  \texttt{salvaruehling@gmail.com} \\
  \And
    Venkatesh Ramesh{$^\ast$} \\
   Universit\'e de Montr\'eal \& Mila\\
  \texttt{venkatesh.ramesh@umontreal.ca} \\
  \And
    Jason N.~S.~Cole\quad Howard Barker \\
    Environment and Climate Change Canada \\
    \texttt{\{jason.cole,howard.barker\}@canada.ca}
    \And
    David Rolnick \\
    McGill University \& Mila \\
    \texttt{drolnick@cs.mcgill.ca}
}
\begin{document}

\maketitle
% ---------------------------------------- Main paper starts
\input{_packages_macros/macros}

\input{sections/0abstract}
\input{sections/1intro}

\input{figures/latex/climatechange}
\input{sections/2relatedWork}

\input{sections/2zbackground}

\input{sections/3dataset}

\input{sections/4experiments}

\input{sections/6extensions}

% ---------------------------------------- Main paper ends
\newpage
% ACKNOWLEDGMENTS
\section*{Acknowledgements}
This research was in part supported by a Canada CIFAR AI Chair and 
funding has been provided by Environment and Climate Change Canada.
\bibliography{references}
\bibliographystyle{plainnat} %ieeetran, plain
\input{appendix/_main}

\end{document}

%% file: _packages_macros/table_pckgs.tex
\usepackage{xcolor}
\usepackage{multirow}
\usepackage{colortbl}
\usepackage{makecell}
\usepackage{caption}
\usepackage{arydshln}
\usepackage{pifont}
\usepackage{siunitx}
\usepackage{booktabs}       % professional-quality tables

\definecolor{cmark-color}{HTML}{0080ff}  % 27932B
\definecolor{xmark-color}{HTML}{A0A0A0}  % D81F1F

\definecolor{white}{HTML}{FFFFFF}
\definecolor{lightgray}{HTML}{F3F3F3}
\definecolor{black}{HTML}{000000}
\definecolor{tabletagcolor}{HTML}{BDBDBD}

\definecolor{cell}{HTML}{B0BEC5}
\definecolor{rowbackground}{HTML}{F0F2F4}

\newcommand{\cmark}{\Large{\textcolor{cmark-color}{\checkmark}}}

%% file: _packages_macros/macros.tex
\newcommand{\E}{\mathbb{E}} % expected value
\newcommand{\R}{\mathbb{R}} % real numbers

%%% FOR BENCHMARK HIGHLIGHTING:
\newcommand{\embf}[1]{\emph{\textbf{{#1}}}}
\newcommand{\red}[1]{{\color{red}{#1}}}
\newcommand{\blue}[1]{{\color{blue}{#1}}}
\newcommand{\edited}[1]{{#1}}
\newcommand{\first}[1]{{\textbf{#1}}}
\newcommand{\third}[1]{{\emph{#1}}}
\newcommand{\Abl}[2]{{#1 $\pm$ #2}}
\newcommand{\emphAbl}[2]{{\textbf{#1 $\pm$ #2}}}
\newcommand{\emphBadAbl}[2]{{\color{red}\textbf{#1 $\pm$ #2}}}

\newcommand{\climart}{ClimART}
\newcommand{\bfclimart}{\textbf{\climart}}

\newcommand*{\todoECCC}{\textbf{\textcolor{red}{\\TODO ECCC}}}
\newcommand{\globals}{\emph{global}}
\newcommand{\levels}{\emph{levels}}
\newcommand{\layers}{\emph{\layers}}

%% file: sections/0abstract.tex
\begin{abstract}
  Numerical simulations of Earth's weather and climate require substantial amounts of computation. This has led to a growing interest in replacing subroutines that explicitly compute physical processes with approximate machine learning (ML) methods that are fast at inference time. Within weather and climate models, atmospheric radiative transfer (RT) calculations are especially expensive. 
  This has made them a popular target for neural network-based emulators.  
  However, prior work is hard to compare due to the lack of a comprehensive dataset and standardized best practices for ML benchmarking.
  To fill this gap, we build a large dataset, \climart, with more than \emph{10 million samples from present, pre-industrial, and future climate conditions}, based on the Canadian Earth System Model.
  \climart\ poses several methodological challenges for the ML community, such as multiple out-of-distribution test sets, underlying domain physics, and a trade-off between accuracy and inference speed.
  We also present several novel baselines that indicate shortcomings of datasets and network architectures used in prior work.\footnote{\edited{Download instructions, baselines, and code are available at: \url{https://github.com/RolnickLab/climart}}}
\end{abstract}

%% file: sections/1intro.tex
\section{Introduction}
\label{sec:intro}
Numerical weather prediction (NWP) models have become essential tools for numerous sectors of society. Their close relatives, global and regional climate models (GRCM) provide crucial information to policymakers and the public about Earth's changing climate and its various impacts on the biosphere.
These models attempt to simulate many complicated physical processes that interact over wide ranges of space and time and seamlessly link Earth's atmosphere, ocean, land, and ice.
However, due to the complexity and number of physical processes that have to be addressed, various simplifications must in practice be made, involving mathematical and numerical approximations that often have substantial statistical bias and computational cost.

One of these approximations is the {\it sub-grid scale parametrization} that is routinely used to approximate atmospheric radiative transfer (RT). 
The RT routine has traditionally represented the largest computational bottleneck in most weather and climate simulation. To speed up the computation, this routine is run only every few iterations and the results for the intermediate iterations have to be interpolated. Needless to say, this approximation of intermediate steps introduces errors in climate and weather predictions. The computational cost also means that climate models must be run at extremely coarse spatial resolutions that leave most processes unresolved and reduce the utility of information in predicting and responding to the effects of climate change.

Thanks to their better inference speed, neural network-based \emph{surrogates} are a promising alternative to computationally slow physics parametrizations. 
Such hybrid modelling approaches, however, present several challenges, including accurate emulation of complex physical processes, as well as the (out-of-distribution) {\it generalization power} of ML models to handle environmental conditions not present in their training datasets (e.g., weather states that are not realized under current conditions).

To date, different datasets, setups, and evaluation procedures have made results in this space hard to compare. 
This can be attributed to the fact that no comprehensive public dataset exists, and the creation of it requires access to, and knowledge of, the relevant climate model. 
To address these issues and catalyze further work, we introduce a new and comprehensive dataset for the RT problem and open-source it under the Creative Commons license.

Our key contributions include: 

\begin{itemize}
    \item \textbf{\climart : Climate Atmospheric Radiative Transfer}, is the \emph{most comprehensive} publicly available dataset for ML emulation of weather and climate model parameterizations. It comes with \emph{more than 10 million samples}, including three subsets of data for evaluating \emph{out-of-distribution (OOD) generalization}. 
    \item
    \textbf{Applying New Models to ClimART}. We propose multiple new models not studied in the related work, which thanks to the comprehensiveness of ClimART, allow us to \emph{analyze the limitations of previously used models and datasets}.
    \item
    \textbf{Towards Advancing the State-of-the-Art}.
    \climart's scale, unique properties, and ease of access, together with the accompanying code interface and baselines, will lower barriers for the ML community to tackle impactful challenges in climate science. 
    \climart~also presents opportunities for spurring methodological innovation in ML via multiple \emph{out-of-distribution test sets}, the scope for building \emph{physics-informed ML models}, and the \emph{accuracy versus inference speed trade-off} inherent in the problem setting.
    
\end{itemize}

%% file: figures/latex/climatechange.tex
\begin{figure}[h]%
    \centering
    \subfloat[\centering Surface Temperature on Jan.~1, 1979]{{\includegraphics[scale=0.28]{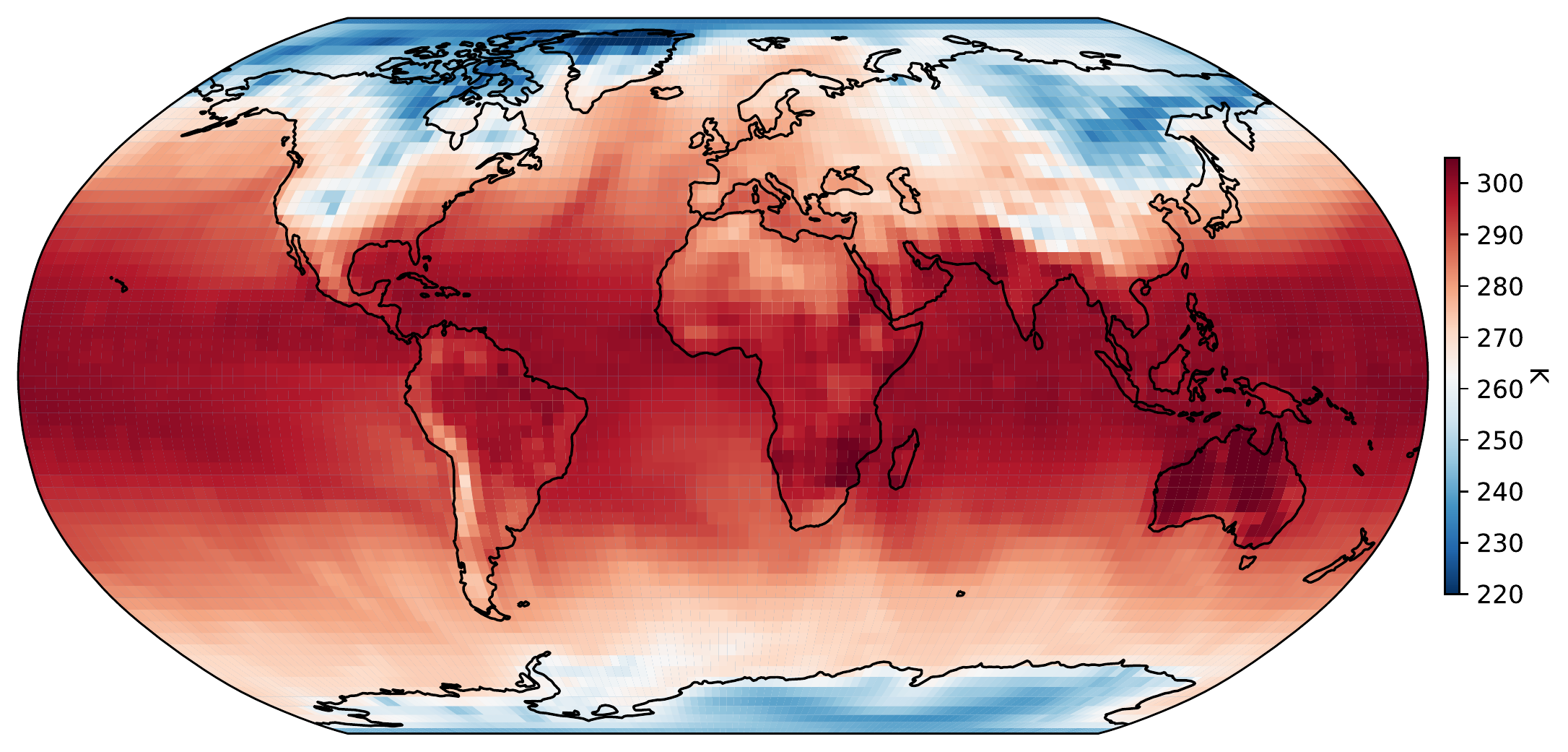} }}%
    \qquad
    \subfloat[\centering Projected Surface Temperature on Jan.~1, 2070]
    {{\includegraphics[scale=0.28]{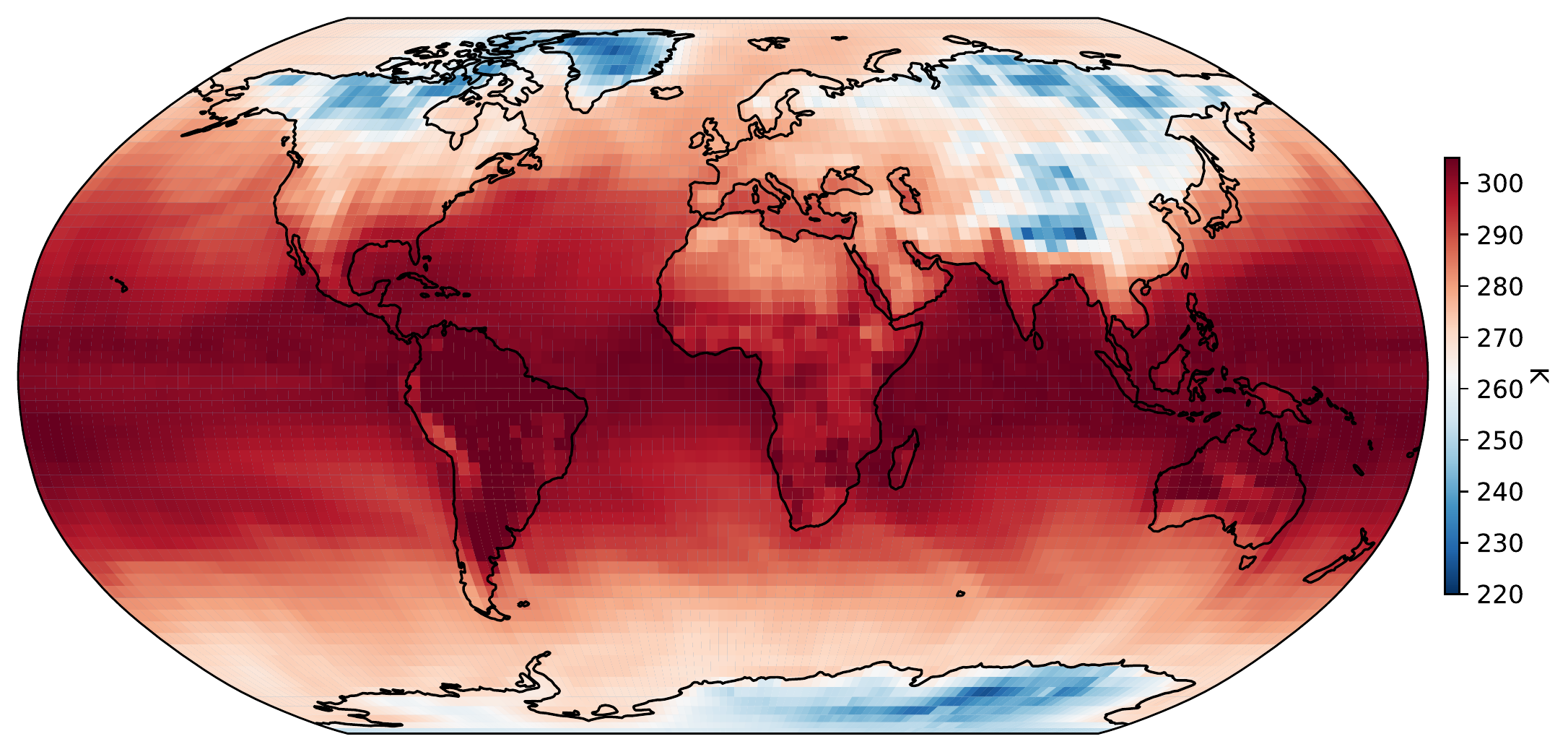} }}%
    \caption{Surface temperatures in atmospheric snapshots from 1979 and 2070.}%
    \label{fig:example}%
\end{figure}

%% file: sections/2relatedWork.tex
\section{Related Work}
\label{sec:relatedWord}
\input{figures/latex/model_per_all}
There have been various recent efforts to emulate sub-grid scale parameterizations by neural networks, which, thanks to their computational efficiencies, are expected to significantly speed up large-scale model simulations \cite{Brenowitz2018, Rasp2018, yuval2020stable}. 

\edited{The computational burden of the RT physics motivated early pioneering work to seek out its emulation with shallow multi-layer perceptron (MLP) networks \cite{cheruy1996RTML, chevallier1998RTML,  krasnoNcep2010, krasnopolsky2005new}, including decadal climate model simulations \cite{krasnoDecadalRTMLsim2008}.}
More recent work \edited{still focuses on using MLPs to emulate (a part of) the} RT physics \cite{radnet2020, 3d_RT_ML2021, palE3SM2019, opticsRTML2020, opticsRTML2021}. \edited{2D CNNs have been also used in \cite{radnet2020}, which however treat the different input variables within the second spatial dimension instead of in the channel dimension.}
Prior work on such ML emulators, however, employed datasets that simplify Earth (e.g., with Aqua-planet conditions \cite{Brenowitz2018, Rasp2018, yuval2020stable}), use a limited subset of climate model variables as predictors~\cite{krasnoNcep2010, radnet2020, Rasp2018, opticsRTML2021},
use manually perturbed test sets~\cite{radnet2020, opticsRTML2021}, and generally fail to accurately probe the generalization power of ML models ~\cite{radnet2020, 3d_RT_ML2021, palE3SM2019, opticsRTML2021}.
% Radiative Forcing Model Intercomparison Project} (RFMIP)
The latter is particularly important, randomly-split test sets~\cite{3d_RT_ML2021, palE3SM2019, opticsRTML2021}, and/or test data coming from at most two different years~\cite{radnet2020, 3d_RT_ML2021, palE3SM2019} can overestimate the actual skill of the ML emulators on real world unseen data. This can be fatal when the ML emulator is to be used in long-term simulations (e.g., future climate projections).

In physics-based RT models as used in large-scale environmental models, computation of radiative flux profiles is a two-step process. The first step involves the calculation of optical properties for gases, aerosols, and clouds. The second, more computationally intensive, and arguably most erroneous step is the application of a solution of the RT equation, using the optical properties from the first step, leading to vertical profiles of radiative fluxes. Work by \cite{opticsRTML2020, opticsRTML2021} focused on a hybrid approach in which gaseous optical properties are predicted by an ML model and then passed off to a physics-based RT model for calculation of fluxes. However, \cite{opticsRTML2021} generates training data by just using perturbations on input variables of RFMIP\cite{RFMIP}. \cite{opticsRTML2020} make a more comprehensive effort via generation of training data by combining multiple sources such as RFMIP\cite{RFMIP}, CKDMIP\cite{CKDMIP} and CAMS\cite{CAMS}, and perturbing them to generate sufficient training data. Such perturbations, however, might lead to unrealistic input values.

The present work is closer to \cite{radnet2020} that used a small fraction of ERA-Interim data from 1979-85 and 2015-16.
Their concern was however, limited, to longwave radiative flux profiles for simplified clear-sky atmospheric conditions (without greenhouse gases like methane). Similarly to \cite{opticsRTML2021}, they employ a test set that includes manually perturbed atmospheric states. We believe that our dataset that includes data of pre-industrial and future climate drawn from an actual climate model, is more realistic and the better choice.

%% file: figures/latex/model_per_all.tex
\begin{figure}
    \begin{subfigure}{.5\textwidth}
      \centering
      \includegraphics[width=.95\linewidth]{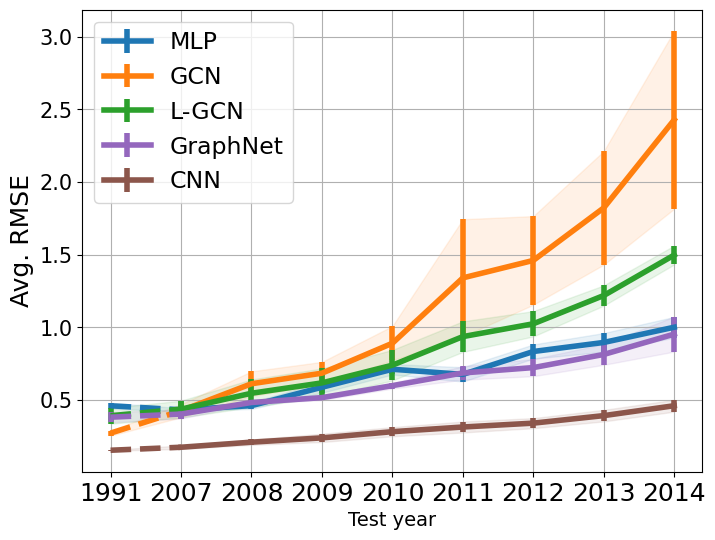}
     \caption{}%Test RMSE}
    \label{fig:yearly_rmse}
    \end{subfigure}%
    \begin{subfigure}{.5\textwidth}
      \centering
      \includegraphics[width=.95\linewidth]{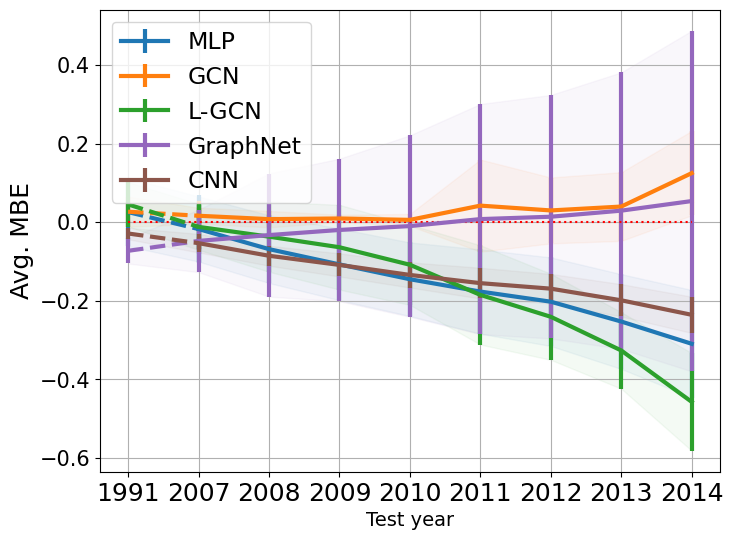}
     \caption{}%Test MBE}
     \label{fig:yearly_mbe}
    \end{subfigure}
    \vspace{-5mm}
    \caption{Present-day test performance}
    \label{fig:yearly}
   \begin{subfigure}{.5\textwidth}
      \centering
      \includegraphics[width=.95\linewidth]{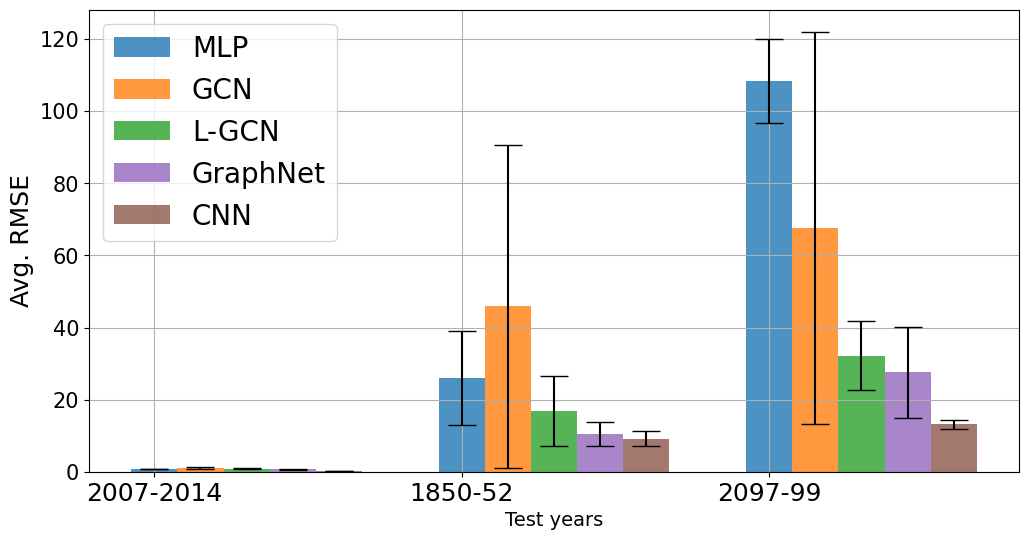}
      \caption{Test RMSE}
      \label{fig:ood_rmse}
    \end{subfigure}%
    \begin{subfigure}{.5\textwidth}
      \centering
      \includegraphics[width=.95\linewidth]{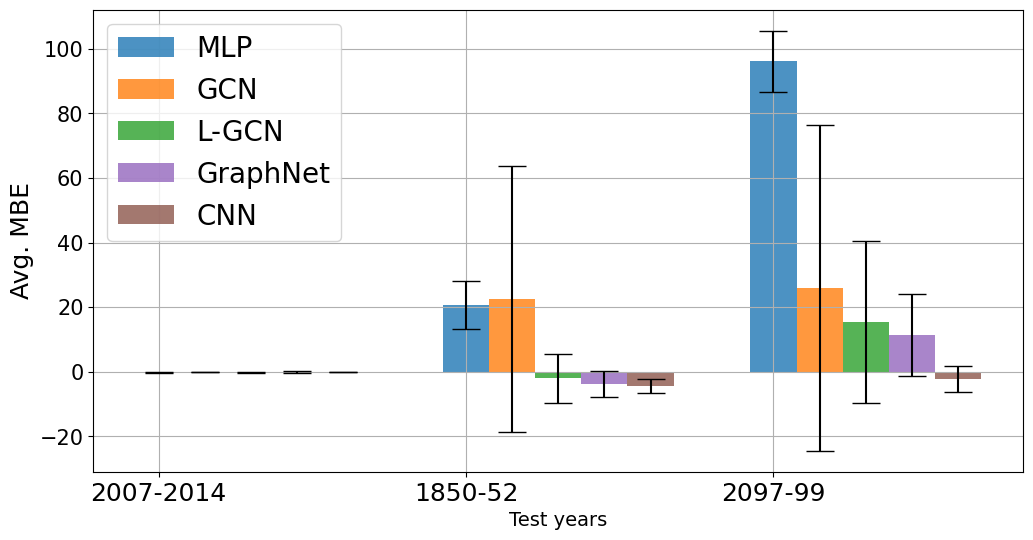}
      \caption{Test MBE}
      \label{fig:ood_mbe}
    \end{subfigure}
    \caption{Performance as a function of the main test set year (Fig. \ref{fig:yearly}) and OOD test set (Fig. \ref{fig:model_per_ood}) for our different baseline models. Metrics are in $W/m^2$ and shown as the average over the vertical and over the up- and down-welling flux errors.
    The leftmost x-tick for the OOD plots corresponds to the average metric of the main test years 2007-2014, for which the metrics are shown on a yearly basis in Fig. \ref{fig:yearly}.
    Generalization to pre-industrial and future pristine-sky conditions is a particularly challenging task because of the changes in gas concentrations 
    More structured models like a CNN or GraphNet perform significantly better than an MLP.
          }
  \label{fig:model_per_ood}
\end{figure}

%% file: sections/2zbackground.tex
\section{Background}
\label{sec:background}
In the following, we introduce background and terminology that is helpful in order to better understand the problem setup and the dataset we present.
\paragraph{Earth System Models}
Earth System Models (ESMs) simulate Earth's climate, including interactions between the atmosphere, ocean, sea-ice and carbon cycle.  Within the atmosphere, the ESM computes the current state of the modelled variables which includes temperature, water vapor, aerosols, and clouds.  This is done using models that discretize the Earth's atmosphere into a 3D spatial grid along latitude, longitude, and vertical dimensions, as well as a discretized timestep.
At each discretized point in the horizontal, one can envision processes occurring in the vertical within a \emph{column} that provides a \emph{profile} of information about the state of the atmosphere, e.g., temperature.  These profiles can be provided with respect to $n$ \emph{layers} in the atmosphere, in addition to $n+1$ \emph{levels} (interfaces between neighboring layers), where the bottom-most level corresponds to the surface and the top-most level corresponds to the top-of-atmosphere (TOA).

\edited{\paragraph{Atmospheric radiative transfer}
Radiative transfer describes the propagation of radiation. In the atmosphere, radiation transmits energy between atmospheric layers. It is classified into \emph{shortwave} (solar), and \emph{longwave} (thermal) radiation, which are emitted by the sun and Earth, respectively. In general, shortwave radiation is absorbed by the Earth and then re-emitted as longwave radiation; some of this longwave radiation is then re-absorbed by the atmosphere. 
At any given time and level of the atmosphere, \emph{up- and down-welling fluxes} refer to the amount of radiation that is traveling up and down between the adjacent atmospheric layers.
As a result of emitting and absorbing radiation, the layers of the atmosphere change temperature, which is captured in \emph{heating rate} profiles (also called cooling rates when negative). The heating rate of any given layer can be directly computed based on the up- and down-welling fluxes of the two adjacent levels (see Appendix \ref{sec:heating_rates}).
Ultimately, the difference in radiative flux into and out of the atmosphere is responsible for changes in the Earth's overall temperature, and the dependence of radiative fluxes on gas concentrations is the source of climate change, since human activity has increased the concentration of various greenhouse gases such as carbon dioxide, methane, and nitrous oxide.

To predict radiative flux and heating rate profiles in the atmosphere, climate scientists work with three types of models for the sky, which differ in the kinds of information they factor in: Pristine, clear, and clouds. \emph{Pristine-sky} is the simplest, meaning that only the concentrations of gases are factored in. \emph{Clear-sky} also includes the concentrations of aerosols, which are particles present in the air such as sulfur-containing compounds. The most general case also includes clouds. 
}

\paragraph{CanESM and its radiative transfer parameterization}
The Canadian Earth System Model (CanESM) is a comprehensive global model used to simulate Earth's past climate and the present results of climate change, as well as to make future climate projections. Figure 1 shows an example of CanESM's simulation of current (1979) and future (2070) surface temperatures.
Its most recent version, CanESM5~\cite{CanESM5}, simulates the atmosphere, ocean, sea-ice, land and carbon cycle, including the coupling between each of these components.  For the atmosphere it uses parameterizations to represent unresolved sub-grid scale processes like radiation, convection, aerosols, and clouds.
\\
The radiative transfer parameterization in CanESM5, is representative of the approach used in most modern ESMs.  The optical properties of a number of components are accounted for, including the surfaces, aerosols, clouds and gases (represented using a correlated $k$-distribution model).
\edited{The parameterization follows an independent column approximation. This intuitively means that for a given latitude and longitude, the RT physics model takes as input only information from the 1D vertical profile of the atmospheric state at the corresponding geographical location.}
More details on CanESM and its RT parametrization can be found in Appendix \ref{sec:appendix_canesm}.

%% file: sections/3dataset.tex
\section{\climart\ Dataset}
\label{sec:dataset}
\subsection{Dataset collection}
For our main dataset, global snapshots \edited{of the current atmospheric state}  were sampled from CanESM5 simulations every $205$ hours from 1979 to 2014.\footnote{The choice of $205$ hours provides a manageable amount of equally spaced data while also ensuring that every hour of the day is covered since 205 is relatively prime to 24.}
CanESM5's horizontal grid discretizes longitude into 128 columns with equal size and latitude into 64 columns using a Gaussian grid (\edited{$8192 = 128\times64$} columns in total).  This results in ~$43$ global snapshots per year for a total of more than 12 million columns for the period 1979-2014 and a raw dataset size of $1.5$TB.
Each column of atmospheric-surface properties was then passed through CanESM5's RT physics model in order to collect the corresponding \edited{RT output: S}hortwave and longwave (up- and down-welling) flux and heating rate profiles for pristine- and clear-sky conditions.
The resulting NetCDF4 datasets were then processed to NumPy arrays stored in Hdf5 format (one per year), with three distinct input arrays as described in the following subsection, and one output array per potential target variable.
We proceeded analogously for the pre-industrial and future climate years, 1850-52 and 2097-99 respectively (see section \ref{sec:appendix_hist_future_ood_collection}).
More details are available in the Appendix \ref{sec:appendix_dset_collection}.

\subsection{Dataset interface}
\paragraph{Inputs}
    We saved the exact same inputs used by the CanESM5 radiation code and augmented them by auxiliary variables such as geographical information (see Appendix \ref{sec:appendix_vars} for full details and description).
    Each input corresponds to a column of CanESM5. Its variables can be divided into three distinct types: i) layer variables, ii) level variables, iii) variables not tied to the height/vertical discretization. Examples for the two first 1D variables are pressure (occurring at both, levels and layers) and water vapour (only present at the layers). The third type of variables are comprised of optical properties of the surface, boundary conditions, and geographical information related data. We refer to this set as the \emph{global} variables. 
    \embf{The data thus has a unique structure with heterogenous data types, where 1D vertical data is complemented by non-spatial information.} We also note that the RT problem is non-local, since the spectral composition, and thus the heating rate, at one level can depend much on attenuation, or production, of radiation at a far-removed layer (e.g., reduced absorption of solar radiation by water vapour near the surface due to the presence of reflective high-altitude cirrus clouds).

\paragraph{Outputs}
    We provide the full radiation output of CanESM5's as a potential target for pristine- and clear-sky conditions.
    That is, \climart\ \embf{comes with two levels of complexity}: pristine-sky (no aerosols and no clouds) being the simpler one compared to clear-sky, which also reflects the impacts on RT due to aerosols.
    It consists, for both shortwave and longwave radiation, of the up-and down-welling flux profiles and corresponding heating rate profiles  (i.e.  $6 = 2\times 3$ distinct variables for each sky condition).

    In our experiments we focus on pristine shortwave radiation. Our dataset, however, allows the user to choose the desired target variables based on their needs.
    
\subsection{Dataset split}
\label{sec:dset_splits}
In the following we describe the data split that we recommend to follow for benchmarking purposes. 
\vspace{-2mm}
\paragraph{Training and Validation sets}
\climart\ provides the complete data extracted from CanESM5, as described above, from 1979 to 2006, excluding 1991-93, as suggested data for training and validating ML models. In our experiments we used 1990, 1999, and 2003 for training, while keeping 2005 for validation.

\paragraph{Main Test set}
We suggest to use the data from the years 2007 to 2014 as main test set. % (as done in all of our experiments).
This relatively long interval allows to test the ML model on a very diverse set of present-day conditions. To make evaluations feasible regardless of the available compute resources, we chose to subsample 15 random snapshots for each year. This results in almost 1 million testing samples.

\subsubsection{Out-of-distribution (OOD) test sets}
    In order to evaluate how well a ML model generalizes beyond the present-day conditions found in the main dataset, we provide \emph{three distinct OOD test sets} \edited{that cover an anomaly in the atmospheric state, as well as two temporal distributional shifts.}

\paragraph{Mount Pinatubo eruption}

\edited{This test set includes} conditions from the year 1991, when the Mount Pinatubo volcano erupted, and probes how well the ML model can cope with sudden atmospheric changes.
The challenges arise via a sudden increase in atmospheric opacity due to high-altitude volcanic aerosol (see Appendix \ref{sec:appendix_pinatubo} for more details). While CanESM5's solar RT model deals with these aerosols well, it is the specification of changes in aerosol mass and distribution (i.e., inputs to the RT model) that pose the largest challenges.
\edited{
Since remnants of the emitted stratospheric aerosols remained in the atmosphere for years after the eruption, we chose to exclude the two subsequent years, 1992 and 1993, from \climart\ in order to avoid data leakage during training.
}
\paragraph{Pre-industrial and future}
\label{sec:appendix_hist_future_ood_collection}

\edited{
This test set probes how well the ML emulator generalizes under challenging distributional shifts. For this purpose, \climart\ provides historic data from the years 1850-52 and future data from the years 2097-99.} In both cases, the primary challenge for ML models is that they can be expected to encounter surface-atmosphere conditions that are not present in the training dataset.
\edited{The primary differences between current and \emph{pre-industrial} conditions involve: reduced atmospheric trace (greenhouse) gas concentrations for pre-industrial times; changes in aerosol emission; and some land surface properties that arise through changes in land usage. 
The \emph{future climate} data can test how well ML models extrapolate to conditions that differ from the current climatic state as a result of radiative forcing through increases in greenhouse gas concentrations.
Future climatic conditions were simulated by CanESM5 based on increases in atmospheric greenhouse gas concentrations and changes in aerosol emissions that follow well-defined scenarios (see~\cite{CanESM5}) laid out for the Sixth Coupled Model Intercomparison Project (CMIP6; cf. \cite{CMIP6}).
}

\subsection{Usage}
It is important to note that a ML model trained on our dataset be both {\it verified} and {\it validated} before it can be employed "operational" in a global climate or NWP model. The {\it verification} phase  is characterized by "simple" quantitative assessment of a ML model's bias and random (conditional) errors. Once one feels that the ML model is ready to go into the dynamical model, its computation-saving aspect can be assessed against any ramifications it has on the overall forecast of the model. Ideally, a successful ML model (i.e., one that is {\it validated}) will simultaneously reduce computation time and have incur only statistically insignificant impacts on the overall forecast.

Furthermore, we note that the level of "success" of an ML model at estimation of radiative flux profiles can differ for weather and climate applications, for it is expected that these areas of application will tolerate bias and random errors differently. For instance, an ML model with minor, but non-negligible, bias error might be acceptable for short-range weather predictions, but could have untenable affects on longer-term climate projections. Likewise, an ML model's random errors might tend to wash-out in long, low-resolution climate simulations, but might initiate spurious extreme events in high-resolution weather forecasts.

\subsection{Limitations}
\label{sec:limitations}
Firstly, while advances on this problem would directly benefit the whole community, the inconsistent interfaces between different climate models and their parameterizations would likely require re-training the models for those specific input-output interfaces.
\edited{We note that one motivation for proposing fully convolutional and graph-based networks in our experiments, is their applicability regardless of the vertical discretization of the columns (depending on the climate model, a column might be divided into different layers). The shortcoming of MLPs, which do not enjoy this property, was also identified by \cite{opticsRTML2021}.
Applying fully convolutional and graph-based networks to emulate parameterizations with different vertical discretization than the one trained on is an interesting direction for future work and could present a way to have ML emulators that are more generally applicable.}

Secondly, our targets do not include radiation output under all-sky conditions (which, besides aerosols, includes clouds).
We believe however that our otherwise comprehensive dataset will serve well as a test-bed for ML emulators under the more simple (yet complex) pristine- and clear-sky conditions. Moreover, we note that pristine- and clear-sky are routinely used in diagnostic analyses of climate and weather model results. That is, while such conditions do not often occur in the atmosphere (mostly above the troposphere), they are nevertheless computed for the entire globe in order to assess a model’s cloud dynamics and compare to satellite data.

%% file: sections/4experiments.tex
\input{tables/main_ml_benchmark}
\section{Experiments} % Or 
\label{sec:experiments}

\subsection{Benchmarking neural network architectures}
\label{sec:ml_benchmark}
Note that prior work usually restricted the ML model to be a multi-layer perceptron (MLP).
In light of the structured data in \climart, we aim to 1) propose more structured neural network architectures that we believe are more suitable to the task and on which we hope follow-up work can build upon; 2) study how these more structured neural network architectures compare to the unstructured MLP.
\\
Thus, we benchmark an MLP against a 1-D convolutional neural network (CNN)~\cite{CNN}, a graph convolutional network (GCN)~\cite{kipf2017gcn}, and a graph network \cite{battaglia2018graphnet}.
We now give a high-level overview over each of the architectures\edited{, which all are relatively lightweight as it is important to keep the inherent \emph{inference speed versus accuracy trade-off} in mind.} More details on it and used hyperparameters can be found in Appendix \ref{sec:appendix_exps}.
\begin{itemize}
    \item 
    \textbf{MLP:} The MLP used for our experiments is a simple three layer MLP with the following hidden-layer dimensions: $\langle512,256,256\rangle$. As an MLP takes unstructured 1D data as input, all the input variables need to be flattened into a single vector for the MLP.
    \item 
    \textbf{GCNs}, take graph-structured data as input. 
    To map the columns to a graph, we use a straighforward line-graph structure where each node is a level or layer and is connected to the two layers or levels spatially adjacent to it above and below. To take into account the \emph{global} information, we add it as an additional node to the graph with connections to all other nodes. The resulting graph structure, in form of an adjacency matrix, is shown in Fig.\ref{fig:linegraph_adj}, where the global node has index 0, and the other nodes are spatially indexed for plotting purposes, where 1 corresponds to the TOA level and the last node corresponds to the surface. A more sophisticated graph structure is studied in section \ref{sec:non_locality_exps} (\textbf{L-GCN}). 
    \\
    The used GCN has three layers of dimension 128.
    \item 
    \textbf{Graph networks}, take graph-structured data (with node and edge features), complemented by a so-called global feature vector, as input. Thus, it is the most natural model for our task, since we can map the levels to be the nodes, layers to be the edges connecting the adjacent levels (i.e.~a line-graph), and the non-spatial variables to the global feature vector. 
    Essentially, a graph network \cite{battaglia2018graphnet} consists of multiple MLP modules, for which we use 1-layer MLPs with a hidden dimension of 128. We use a three-layered graph network.
    \item 
    \textbf{1D CNN:} For the CNN model, we  use a 3-layer network with kernel sizes $\langle20, 10, 5\rangle$ and the corresponding strides set as $\langle2, 2, 1\rangle$. The channels parameter is given by $\langle200, 400, 100\rangle$, with the last channels setting it equal to the input size. We then apply a global average over the resulting tensor to get the output. To preprocess the data for CNN, we pad the surface and layers variable to match the dimensions of levels variable. Then the result is concatenated and fed to the model.

\end{itemize}

For all the models, we use a learning rate of $2\text{e-}4$ with an exponential decay learning rate scheduler and Adam \cite{kingma2017adam} as optimizer. All the models are trained for 100 epochs with the mean squared error loss. We report root mean squared error (RMSE) and mean bias error (MBE) statistics over three random seeds. 
The results for our baseline models on pristine-sky conditions are shown in Fig.~\ref{fig:yearly} (more detailed in Fig.~\ref{fig:yearly_long}) and reported in Table \ref{tab:ml_benchmark}.
\edited{Notably all models can be seen to significantly deteriorate as a function of the test year (that becomes temporally farther away from the training years). We note that prior work could not observe this phenomenon since the test data did not cover as many years or was randomly split from the training set.}
We also find that the CNN architecture provides the most skillful emulation in terms of RMSE, while the GraphNet provides the least biased errors.
This holds for the respective metrics computed at the TOA and surface level only, as well as when averaged out vertically over all levels.
We note that the surface and TOA flux predictions are especially important, as they are directly used by the host climate or weather model.
We also find that the L-GCN, which extends the GCN by a learnable edge structure module, is able to significantly outperform it, especially for TOA predictions, see next subsection.

\subsection{Exploiting non-locality}
\label{sec:non_locality_exps}
\input{figures/latex/gcn_learned_adj}
Heating and cooling at one layer of the atmosphere depends on attenuation of radiation in all other layers. 
This non-locality complicates numerical simulation of the process greatly, and takes sizable amounts of computer resources to handle properly (see Appendix \ref{sec:appendix_canesm} for more details). Therefore, we expect ML models that can take non-locality into account to be a promising research direction.
In the following we support this hypothesis with a GCN that \emph{learns the edge structure} (i.e. the adjacency matrix of the underlying graph) as proposed in \cite{cachay2021world}, denoted L-GCN. The model architecture and all other parameters are identical to the GCN.
Making the connections between arbitrary layers and levels learnable relaxes the hard-coded inductive bias imposed by the highly local line-graph structure used in the standard GCN model.
Indeed, not only does L-GCN outperform the GCN (Table \ref{tab:ml_benchmark}), but our post-analysis also reveals that it 1) \embf{learns a graph structure very different to the line-graph} used by the GCN and GraphNet (see Fig.~\ref{fig:lgcn_adj}), 2) gives high importance to the \globals\ node, which is expected given the importance of the boundary conditions and surface type of a column (see Appendix \ref{sec:appendix_lgcn} for an analysis based on the eigenvector centrality).

\subsection{Speed}
To assess the speed of our models at inference time, we speed-test the ML models for pristine-sky/clear-sky conditions on CPU and GPU, and speed-test the physics-based models for pristine-sky/clear-sky conditions for different numbers of CPUs.
Note that we did not optimize the forward pass of the ML models for efficiency; thus, greater speed should be readily attainable.

\input{tables/speed/ml_pristine}

For the evaluation of ML models, we make use of an instance with 4 CPUs (\textit{2x AMD EPYC Zen 2 "Rome" 7742}), 12GB Memory and Nvidia v100 GPU. The results for different ML models for pristine-sky conditions is shown in \ref{tab:ml_pristine} 
excluding the time for data loading. These results are averaged over 10 forward passes for an entire snapshot (8192 columns) excluding the first two warm-up passes. As expected, the MLP is fastest for both clear-sky and and pristine-sky inputs. 
\edited{Especially on a GPU the MLP provides, together with the CNN and GraphNet. a considerable speed-up over the RT physics.
This is promising since GPUs are starting to be natively supported within the compute environments in which NWPs and GRCMs run \cite{bauer2021digital}, including CanESM's.}
When evaluated with a batch-size of 8192, the model performs $\textbf{3.5x}$ better than with a batch-size of 512.

The physics-based RT parameterization is not GPU compatible yet so we run it by increasing the number of CPUs in the instance and average them over three runs. The RT parameterization can be significantly sped-up by increasing the number of CPUs from 4 to 16. However, the going from 16 to 64 CPUs has diminishing returns. It should be noted that this physics-based model was run in \emph{offline} mode, where the computation is done serially. When run together with its host weather or climate model, the predictions occur in parallel.

\subsection{OOD generalization}

For evaluating the generalization of our models in OOD data, we run it on historic data (1850-1852) and future data (2097-2099). These experiments are extremely challenging given the limited size of of training set use for baseline models. Apart from this, in historic (pre-industrial) and future conditions, the values of input variables, especially those relating to the concentrations of gases vary quite a lot. For a model to be able to perform well on this data, it has to have understood the role of gas concentrations in prediction of the flux properties. As seen from the results in Fig. \ref{fig:model_per_ood}, all models degrade significantly in performance, especially for future climate conditions. However, it is notable how the models that better account for the structure of atmospheric data perform considerably better compared to MLPs: While both, the MLP and GraphNet, perform comparably well for present-day conditions with an RMSE of less than 1 $W/m^2$ (Fig. \ref{fig:avg_rmse}), the MLP's RMSE for future conditions is above 100 while the GraphNet's stays at around 30 (Fig.~\ref{fig:ood_avg_rmse}). Similarly, the CNN degrades ``only'' from less than 0.5 RMSE on the main test set, to below 18 $W/m^2$ for future-day climate conditions-

%% file: tables/main_ml_benchmark.tex
\begin{table*}[t]
\setlength{\tabcolsep}{5pt}
\renewcommand{\arraystretch}{1.2}
\centering
\small
 \begin{tabular}{p{1.25cm}rrrrrrrrrrrrrr}
 \toprule
 & \multicolumn{2}{c}{\textbf{Vertical Avg.}} & \multicolumn{2}{c}{\textbf{Surface}}
 & \multicolumn{2}{c}{\textbf{TOA}} \\
\cmidrule(l){2-3} \cmidrule(l){4-5}
\cmidrule(l){6-7} 
 \textbf{Model} & RMSE & MBE & RMSE & MBE & RMSE & MBE \\ 
 \midrule
 
MLP	 & 0.701 $\pm$ 0.04 & -0.160 $\pm$ 0.10 & 0.684 $\pm$ 0.07 & -0.282 $\pm$ 0.09 & 0.573 $\pm$ 0.06 & -0.208 $\pm$ 0.10 \\ 
\rowcolor{rowbackground}
GCN	 & 1.209 $\pm$ 0.25 & 0.034 $\pm$ 0.04 & 1.260 $\pm$ 0.70 & 0.244 $\pm$ 0.47 & 0.815 $\pm$ 0.12 & 0.071 $\pm$ 0.15 \\ 
L-GCN	 & 0.878 $\pm$ 0.09 & -0.179 $\pm$ 0.10 & 0.714 $\pm$ 0.21 & -0.241 $\pm$ 0.29 & 0.440 $\pm$ 0.09 & -0.048 $\pm$ 0.21 \\ 
\rowcolor{rowbackground}
GraphNet	 & 0.648 $\pm$ 0.04 & \first{-0.001 $\pm$ 0.25} & 0.620 $\pm$ 0.08 & \first{0.017 $\pm$ 0.29} & 0.434 $\pm$ 0.08 & \first{-0.033 $\pm$ 0.21} \\ 
CNN	 & \first{0.303 $\pm$ 0.03} & -0.142 $\pm$ 0.03 & 
    \first{0.265 $\pm$ 0.03} & -0.138 $\pm$ 0.03 &
    \first{0.284 $\pm$ 0.05} & -0.177 $\pm$ 0.04 \\ 
\bottomrule
\end{tabular}
\caption{
We run several neural network architectures on \climart\ to emulate the shortwave down- and up-welling radiative fluxes. The reported metrics (in $W/m^2$) are averaged out over all test samples (years 2007-2014), three random seeds, as well as over the two errors for up- and down-welling fluxes.
Vertical Avg. also averages the metrics over all levels (heights), see \ref{sec:ml_benchmark} for more.
}
\label{tab:ml_benchmark}
\end{table*}

%% file: figures/latex/gcn_learned_adj.tex
\begin{figure}
    \begin{subfigure}{.5\textwidth}
      \centering
      \includegraphics[width=.95\linewidth]{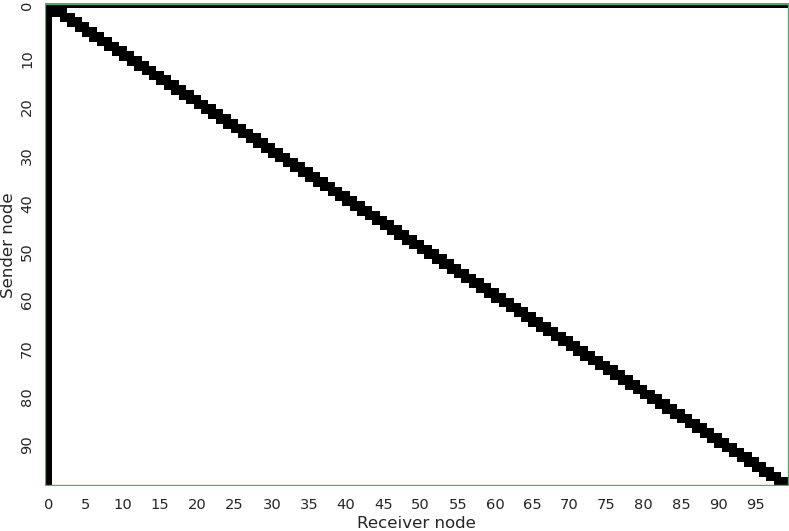}
      \caption{Hard-coded structure for GCN and graph-net}
      %\caption{Adjacency matrix of a line-graph with global node}
      %\caption{Adjacency matrix of L-GCN at random initialization}
      \label{fig:linegraph_adj}
    \end{subfigure}%
    \begin{subfigure}{.5\textwidth}
      \centering
      \includegraphics[width=.95\linewidth]{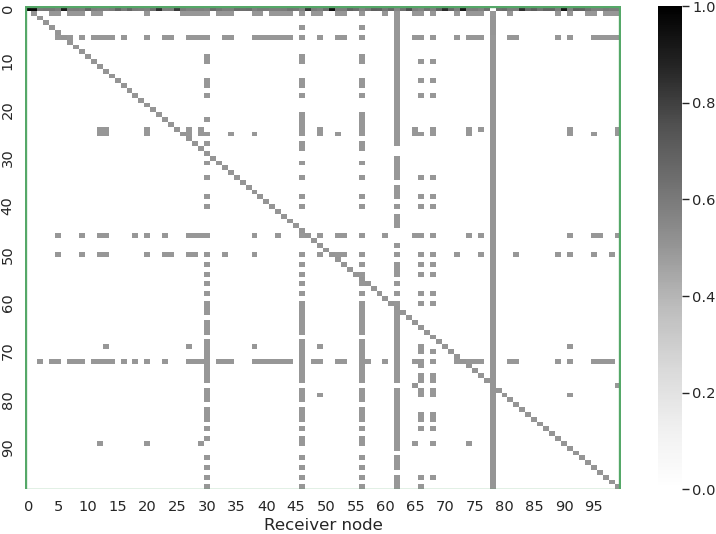}
     \caption{Structure learned by
     L-GCN}
      \label{fig:learned_adj}
    \end{subfigure}
    \caption{L-GCN, a GCN with learnable adjacency matrix, 
    learns an edge structure very different to the diagonal structure of a line-graph (a) that is used is static adjacency matrix for the GCN and graph network baselines. This indicates that the problem benefits from \emph{non-local information}.
    Notably, the many outgoing edges from \globals\ node (index $0$, top row) indicates its importance. 
    Node indices are sorted by height in descending order, i.e index $1$ corresponds to the top-of-atmosphere node/level, and $100$ to the surface level.}
  \label{fig:lgcn_adj}
\end{figure}

%% file: tables/speed/ml_pristine.tex
\begin{table}
\centering
  \begin{tabular}{lcccc}
    \toprule
    \multirow{2}{*}{Model} &
      \multicolumn{2}{c}{Hardware} \\ &
      {CPU} & {GPU} & {Time (s)}\\ 
      \midrule
     \multirow{3}{6em}{Physics-RT} & 2 & $N/A$ & \third{3.3919}\\ 
    & 4 & $N/A$ & 3.3666\\ 
    & 16 & $N/A$ & 2.0988\\ 
    & 64 & $N/A$ & \blue{1.9817}\\ 
    \midrule
    \multirow{2}{6em}{MLP} & 4 & $\times$ & \first{0.1643}\\ 
    & 4 & \checkmark & \first{0.0016}\\
    \midrule
    \multirow{2}{6em}{CNN} & 4 & $\times$ & 3.1870\\ 
    & 4 & \checkmark & \blue{0.0218}\\
    \midrule
    \multirow{2}{6em}{GCN} & 4 & $\times$ & 4.6846\\ 
    & 4 & $\checkmark$ & 1.6818\\
    \midrule
    \multirow{2}{6em}{GraphNet} & 4 & $\times$ & 28.1253\\ 
    & 4 & \checkmark & 0.1659\\
    \bottomrule
  \end{tabular}
\medskip
\caption{\emph{Pristine-sky speed benchmark} of physics-based model and ML models on different hardware configurations. The physics-based model runs serially in offline mode and is not GPU compatible yet. The ML models are all evaluated with a batch size of $8192$ in CPU only and with GPU. The fastest models with and without GPU are in \first{boldened} and the second fastest in \blue{blue}.}
\label{tab:ml_pristine}
\end{table}

%% file: sections/6extensions.tex
\section{Conclusion \& Future Work}
We introduce a novel dataset \climart\ which aims to provide a comprehensive dataset for parameterization of radiative transfer using ML models. We conduct a series of experiments to demonstrate which models are able to perform well under the inherent structure of atmospheric data in \nameref{sec:experiments}. 
Future work for improving upon the current baselines could include: 
\begin{itemize}
    \item 
    Improving the model's inference speed via methods like \embf{weight pruning, model compression, or weight quantization}.
    \item Using a \embf{physics-informed neural network} or loss function to predict realistic values in-line with the equations governing radiative transfer.
    \item
    \embf{Multi-task learning} can be explored to emulate both shortwave and longwave fluxes or heating-rates simultaneously.
    \item
    Using Transformer-based architectures for their ability to perform well with arbitrary sequence lengths and incorporation of an \embf{attention mechanism}.
\end{itemize}
On the dataset side, we plan to extend \climart\ to include all-sky data that includes the complexity due to clouds. We hope that \climart\ will advance both fundamental ML methodology and climate science, and catalyze greater involvement of ML researchers in problems relevant to climate change.

%% file: appendix/_main.tex
\newpage
\appendix
\section*{Appendix}

\section{CanESM and radiative transfer details}
\label{sec:appendix_canesm}
\paragraph{The Canadian Earth System Model (CanESM)}
CanESM is a comprehensive global model used to simulate Earth's climate past and present climate change as well as to make future climate projections.  The most recent version of CanESM is version 5 \cite{CanESM5}.  CanESM5 simulates the atmosphere, ocean, sea-ice, land and carbon cycle, including the interactions between each of these components.  The atmospheric component of CanESM5 is version 5 of the Canadian Atmospheric Model (CanAM5), which simulates a range of atmospheric physical processes, including radiation, convection, aerosols and clouds.  CanAM5 uses parameterizations to represent these unresolved sub-gridscale processes, which are similar to those in its predecessor, CanAM4 \cite{CanAM4}.

\paragraph{CanESM5's radiative transfer parameterization}
The radiative transfer parameterization in CanESM5, is representative of the approach used in most modern ESMs.  The optical properties of a number of components are accounted for, including the surfaces, aerosols, clouds and gases (represented using a correlated $k$-distribution model).  The solar and thermal radiative transfer is computed using a 2-stream solution \cite{CanAM4}. The unresolved, subgrid-scale variability of clouds is treated using the Monte Carlo Independent Column Approximation (McICA) \cite{McICA}.  The subgrid-scale variability of the surface albedo for solar and emissivity for thermal are accounted for in the radiative transfer calculations \cite{CanESM5}.  The performance of the CanESM radiative transfer code under pristine (gas-only), clear (gas plus aerosols) and all-sky conditions has been documented relative to line-by-line calculations and other radiative transfer models with similar complexity \cite{CIRC,protoRFMIP}.

\paragraph{Non-locality in radiative transfer}
While RT models used in large-scale models assume, with some justification, that radiation does not flow laterally between columns, they absolutely have to consider flows of radiation vertically. This means that heating and cooling at one layer depends on attenuation of radiation in all other layers. 
This non-locality complicates numerical simulation of the process greatly, and takes sizable amounts of computer resources to handle properly.
Since the simplifying assumption of horizontally independent columns is expected to be employed for some time still, the hope is that the ML community can successfully apply or develop novel models that adequately handle the vertical non-local aspects of computing atmospheric RT.

\section{Dataset details}
\label{appendix_dataset}

\subsection{Dataset collection}
\label{sec:appendix_dset_collection}
Our dataset focuses on pristine-sky (no aerosols and no clouds) as well as clear-sky (no clouds) conditions, i.e. it leaves out the most general all-sky condition that includes clouds.
These input conditions, which consist of surface properties and profiles of pressure, temperature, humidity, and trace gases, were simulated by setting input variables corresponding to clouds (and aerosols for pristine-sky) to zero.
These input snapshots, for the respective atmospheric conditions, were then forwarded through CanESM5's RT physics code.
for each atmospheric condition, the outputs are profiles of up- and down-welling fluxes for both, shortwave (solar) and longwave (thermal) radiation, plus their respective heating rates.
These raw inputs and outputs are stored in separate NetCDF4 files for each snapshot. All together (for the main dataset of 1979-2014), they amount to over 1.5Tb of data.

\subsection{Extreme volcanic eruption conditions}
\label{sec:appendix_pinatubo}
Occasionally, a volcanic eruption is large enough to inject material and gases well into the stratosphere. When that happens, the resulting aerosol loading spreads over the globe and can remain suspended for periods of time that are long enough to have measurable impacts on remote sensing data and surface-atmosphere climatic conditions. The overwhelming impact is a reduction of solar radiation absorbed by Earth, and hence a slight, but measurable and attributable, reduction in lower-atmospheric and surface temperatures (with other variables responding according which can both amplify or mitigate the initial cooling). To account for these radiative forcings with some confidence in a global model requires reliable input of height-dependent mass loading and aerosol optical properties. If these can be supplied, simple solar RT models can predict accurate flux perturbations. For an ML model to be able to address them well, however, appropriate inputs and responses must be included in the training dataset. The added challenge is that not all volcanoes are equal and their time and location can result in distinct radiative forcings.

\subsection{Complete list and description of variables}
\label{sec:appendix_vars}
A complete list of all input variables can be found in Table \ref{tab:input_vars}, and for all potential target variables in Table \ref{tab:output_vars}. 
\input{appendix/tables/invars}
\input{appendix/tables/outvars}
Within a CanESM grid box, the surface can include multiple types.  An example of this is a grid box that includes a coast line which includes both land and water.  The fraction of the grid box and its optical properties for a particular surface type is passed into the radiative transfer code where it is accounted for in the subsequent calculations.

The variable aerin holds information about the aerosols passed into the radiative transfer calculations.  In the dataset provided these are aerosol mixing ratios.  The third index of the arrays are associated with different aerosols simulated in CanESM5 \cite{CanAM4},
\begin{itemize}
    \item 1: SO4
    \item 2: Accumulation mode sea salt
    \item 3: Coarse mode sea salt
    \item 4: Accumulation mode dust
    \item 5: Coarse mode dust
    \item 6: Hydrophobic black carbon
    \item 7: Hydrophyllic black carbon
    \item 8: Hydrophobic organic carbon
    \item 9: Hydrophyllic organic carbon
\end{itemize}

\subsection{Computing heating rates based on radiative fluxes}
\label{sec:heating_rates}
The heating rate $h_l$ of any given layer $l \in \{1, ..,S_{lay}\}$ can be directly computed based on the up- and down-welling fluxes of the two adjacent levels as follows:
\begin{equation}
    h_l = c \cdot \frac{(F^\text{up}_{l+1} - F^\text{down}_{l+1}) - (F^\text{up}_{l} - F^\text{down}_{l})}{p^\text{lev}_{l+1} - p^\text{lev}_l},
\end{equation}
where $c \approx 9.76\text{e}^{-3}$, and $F^\text{up}_k, F^\text{down}_k$, and $p^\text{lev}_k$ are the corresponding up-welling flux, down-welling flux, and pressure, of a level $k\in \{1, .., S_{lay}+1\}$.

\subsection{Dataset interface}
\subsubsection{Inputs pre-processing}
\label{sec:inputs_preprocessing}
To decrease the dataset size as well as mapping the raw variables into a format that is more amenable for ML models, we chose to concatenate the input variables that share the same spatial dimension across the feature/channel dimension.
The information about which channel corresponds to which variable was saved, and is provided in the \texttt{META\_INFO.json} file included in the dataset root directory.
This results in three distinct input types and arrays per sample:
\begin{itemize}
    \item \emph{globals}: Consists of variables related to boundary conditions (e.g. sun angle), surface type variables, as well as geographical information (as described in \ref{sec:appendix_geo_info}), which all do not have a spatial dimension (i.e. one, possibly multi-dimensional, variable per column/data example).
    \item \emph{levels}: Consists of variables occurring at each level of the column (50 levels in this case). These are only four variables. It is worth recalling that the target radiative flux profiles are level variables.
     \item \emph{layers}: Consists of variables occurring at each layer of the column (49 layers in this case). It is worth recalling that the target heating rate profiles are layer variables (although they can be computed based on the up- and down-welling fluxes).
\end{itemize}
There are 82 global features per column, 4 level features per level, and 14 (45 for clear-sky) layer features per layer. 
Thus, all together there are a total of $2487 = 82 + 4 \times 50 + 45 \times 49$ (968 for pristine-sky) potential features.

\subsubsection{Data normalization}
For convenience, we provide pre-computed dataset statistics (mean, standard deviation, minimum and maximum) in the \texttt{statistics.npz} file that can be found in the root directory of the dataset.
All statistics were computed on 1979-1990 + 1994-2004, i.e. on the years that we propose to use for training. 
Given the large sample size, it is important to use float64 precision for the mean and standard deviation in order to avoid numerical overflows.
The statistics are provided for each input type, \texttt{in-type} $\in\{$\emph{layers}, \emph{levels}, \emph{globals}$\}$, separately and the corresponding arrays have the same feature/channel dimensionality so that they can be directly used for normalization.
The statistics that follow the naming \texttt{<statistic>\_<in-type>} are concatenated scalar statistics for each variable. The statistics that follow the naming \texttt{spatial\_<statistic>\_<in-type>} were, additionally, computed for each level or layer separately (and are thus 2D arrays).
In our experiments we used these statistics to scale the input data to have zero mean and unit standard deviation (\emph{"z-scaling"}), as is common.

\subsubsection{Storing scheme}
\label{sec:storing_scheme}
\paragraph{Inputs}
Recall that each example in \climart\ consists of three distinct input arrays that correspond to the \emph{globals}, \emph{layers}, and \emph{levels} data subset.
All three arrays are stored together in a single Hdf5 file for each year, which can all be found in the \texttt{inputs/} sub-directory.
\\
The \emph{layers} array is concatenated along the channel dimension in such a way, that the 14 first features are the ones needed for pristine-sky experiments, while the whole array would be used for clear-sky experiments.
This avoids storage redundancy, and allows it to access the pristine-sky data by simple slicing of the \emph{layers} array (see \ref{sec:input_dims} for the exact shape of the input arrays).

\paragraph{Outputs}
To allow flexible use of the potential target variables, we store one array per output variable together in a single Hdf5 file per year (a list of all possible target variables is given in Table \ref{tab:output_vars}). Since the targets differ between pristine- and clear-sky conditions, they are stored into the \texttt{outputs\_pristine/} and  \texttt{outputs\_clear\_sky/} sub-directories, respectively.

\paragraph{Directory structure}
The dataset is stored as separate Hdf5 files for each year (filenames follow \emph{<year>.h5}).
From the dataset root directory the structure thus follows:
\begin{itemize}
    \item
    \texttt{META\_INFO.json}
    \item
    \texttt{statistics.npz}
    \item
    \texttt{inputs/}
    \begin{itemize}
        \item 
        1850.h5
        \item
        1851.h5
        \item
        1852.h5
        \item
        1979.h5
        \item
        \dots
        \item
        2014.h5
        \item
        \dots
        \item
        2097.h5
        \item
        2098.h5
        \item
        2099.h5
    \end{itemize}
    \item
    \texttt{outputs\_pristine/}
    \begin{itemize}
        \item         \emph{Same as for inputs}
    \end{itemize}
     \item
    \texttt{outputs\_clear\_sky/}
        \begin{itemize}
        \item         \emph{Same as for inputs}
    \end{itemize}
\end{itemize}

\subsubsection{Inputs dimensions}
\label{sec:input_dims}
For this reason and to avoid storage redundancy, we store one single input array for both pristine- and clear-sky conditions.
The dimensions of \climart's input arrays are:
\begin{itemize}
    \item 
    \emph{layers}: $\quad (N, S_\text{lay}, D_\text{lay})$
    \item 
    \emph{levels}: $\quad (N, S_\text{lev}, D_\text{lev})$
    \item 
    \emph{globals}: $\quad (N, D_\text{glob})$,
\end{itemize}
where $N$ is the data dimensions (i.e. the number of examples of a specific year), $S_\text{lay}$ and $S_\text{lev}$ are the number of layers and levels in a column respectively (49 and 50 in this case), and $D_\text{lay}$, $D_\text{lev}$, $D_\text{glob}$ is the number of features/channels for \emph{layers}, \emph{levels}, \emph{globals} respectively.
For both pristine-sky and clear-sky conditions, we have that $D_\text{lev} = 4$ and $D_\text{glob} = 82$, while $D_\text{lay} = 14$ for pristine-sky, and $D_\text{lay}=45$ for clear-sky conditions (see \ref{sec:inputs_preprocessing} for details on the nature of this).
The array for pristine-sky conditions can be easily accessed by slicing the first 14 features out of the stored array, e.g.: 
\begin{equation}
    \texttt{pristine\_array} = \texttt{layers\_array}[:, :, :14]
\end{equation}

\subsection{Reading the dataset in Python}
Using Python, \climart's input and target arrays can be accessed as follows (for the example year 2007, and assuming that the user wants to predict longwave heating rates under pristine-sky conditions):
\definecolor{codeblue}{rgb}{0.25,0.5,0.5}
\definecolor{codekw}{rgb}{0.85, 0.18, 0.50}
\definecolor{mauve}{rgb}{0.58,0,0.82}
\lstset{
  backgroundcolor=\color{white},
  basicstyle=\fontsize{7.5pt}{7.5pt}\ttfamily\selectfont,
  columns=fullflexible,
  breaklines=true,
  captionpos=b,
  commentstyle=\fontsize{7.5pt}{7.5pt}\color{codeblue},
  stringstyle=\fontsize{7.5pt}{7.5pt}\color{mauve},
  keywordstyle=\fontsize{7.5pt}{7.5pt}\color{codekw},
}
\begin{lstlisting}[language=python, mathescape=true]
# Assume that h5py and numpy are installed and we are in the root data directory.
    import h5py
    import numpy as np
    with h5py.File("inputs/2007.h5", 'r') as h5f:
        X = {
            'layers': np.array(h5f['layers'][..., :14]),  # for clear-sky targets no slicing is needed!
            'levels': np.array(h5f['levels']),
            'globals': np.array(h5f['globals'])
        }
    with h5py.File("outputs_pristine/2007.h5", 'r') as h5f:
        Y = np.array(h5py['hrlc'])  # or take any other variable from Table 4
\end{lstlisting}

\subsection{Dataset split sizes}
Recall that each snapshot (the state of CanESM5 at some timestep) consists of 8192 columns/samples.
Further, recall that by sampling every 205 hours, each year contains either 42 or 43 snapshots ($344064$ or $352,256$ total samples).

We provide the complete data for the years 1979 to 2006, excluding the years 1992-93 in order to avoid potential data leakage when using 1991 as an out-of-distribution test set.
In total there are thus 10,076,160 samples for this period. Thus, minus the held-out year 1991, this results in up to $9,732,096$ potential training samples from present-day conditions.

For the suggested testing period, 2007 to 2014 (inclusive), we randomly subsampled 15 out of the $43$ snapshots per year (giving $122,880$ distinct samples per year). In total this results in $983,040$ samples for the eight testing years.
Randomly subsampling on a yearly basis ensures a diverse test set (as opposed to sampling the snapshots from the same yearly timesteps), which is further magnified by the yearly variability.

\subsection{Adding geographical information}
\label{sec:appendix_geo_info}
Coordinates in the latitude-longitude system are two features used to represent a 3-D space. Due to this, they are not the optimal choice for a ML model to get informed about the three dimensional Earth.
To deal with this issue, we map them to x, y, and z coordinates on a unit sphere. This ensures that the extreme longitudes are close by in the new coordinates.
Concretely, we set for each columns with latitude $lat$ and longitude $lon$ as follows:
%     $x$-$cord = \cos(lat) * \cos(lon) $
\begin{align*}
    x\text{-}cord &= \cos(lat) * \cos(lon)  &&
    y\text{-}cord = \cos(lat) * \sin(lon) &&
    z\text{-}cord = \sin(lat)
\end{align*}

\input{figures/latex/models_per_year_long}
\input{figures/latex/model_per_ood_long}

\section{Experiments}
\label{sec:appendix_exps}
\subsection{Implementation details}
\label{sec:impl_details}
\subsubsection{Model architectures}

\paragraph{MLP}

The MLP used for our experiments is a simple three layer MLP with the following hidden-layer dimensions: $\langle512,256,256\rangle$. As an MLP takes unstructured 1D data as input, all the input variables need to be flattened into a single vector for the MLP.

\paragraph{CNN}

For the CNN model, we  use a 3-layer network with kernel sizes $\langle20, 10, 5\rangle$ and the corresponding strides set as $\langle2, 2, 1\rangle$. The channels parameter is given by $\langle200, 400, 100\rangle$, with the last channels setting it equal to the input size. We then apply a global average pooling over the resulting tensor to get the output. To preprocess the data for CNN, we pad the surface and layers variable to match the dimensions of levels variable. Then the result is concatenated and fed to the model.

\paragraph{Graph Convolutional Network (GCN) and L-GCN}
We use a three-layer GCN \cite{kipf2017gcn} with hidden dimensionality 128 and residual connections.
As nodes of the graph we use all three input types: \emph{layers}, \emph{levels}, and \emph{globals}. The latter is mapped to a global node that is connected to all other nodes, while for the edge structure we use a simple line-graph that contains connections between adjacent levels and layers only.
Thus, the graph has $49 + 50 + 1 = 100$ nodes.
As is standard practice, we add self-loops to the adjacency matrix, see Fig. \ref{fig:linegraph_adj} for a visualization of the resulting adjacency matrix.
Since \emph{layers}, \emph{levels}, and \emph{globals} are heterogenous data arrays with different numbers of features, we project them to a hidden size of 128 with a separate 1-layer MLP for each of the input types, before passing it to the GCN. The MLP projectors use LayerNorm and GeLU as activation function.
The GCN backbone is the same for both GCN and L-GCN, i.e. L-GCN only differs from GCN in its structure learning module, which is identical to the one proposed by \cite{cachay2021world}.
To get predictions we use a 1-layer MLP head that takes as input mean-pooled node embeddings generated by the last GCN layer.
\paragraph{Graph network}
We use a three-layer graph network (GraphNet)\cite{battaglia2018graphnet}, i.e. with three sequential graph network blocks, that do not share weights.
As in \cite{battaglia2018graphnet}, each GraphNet block consists of three update functions for each of the three graph components: global, node, and edge features. The update functions are modeled by distinct 1-layer MLPs with hidden size of 128. Each block uses residual connections.
For the graph structure, we use similarly to the GCN a line-graph with self-loops.
However, a GraphNet enables more modeling flexibility, since we can get rid of the global node in the GCN and instead map it to the global feature vector of a GraphNet. A GraphNet also supports edge features, thus we map the layer features to be edge features (and thus layers be treated as edges between adjacent levels). As nodes of the graph we then use the remaining 50 levels.
Similarly to the GCN, we stack a 1-layer MLP on top of the last GraphNet block to predict the desired number of outputs. The MLP inputs are the mean-pooled node (i.e. levels) representations of the last layer. We choose to pool from the nodes/levels since the target variables -- up- and down-welling flux profiles -- are level variables too.

\subsubsection{Hyperparameters}
Recall from \ref{sec:dset_splits}, that we use the years 1990, 1999, and 2003 for training, while validating on 2005 and testing on the proposed test set years 2007-2014.
For all the models, we normalize the input data by subtracting the mean and dividing by the standard deviation that were computed on the potential training years $\{1979-90, 1994-2004\}$.  The targets are \emph{not} normalized in any form, but directly predicted in their raw form by all models.
The batch size used for training all the models was fixed at $128$. 
All models use the GeLU activation function \cite{hendrycks2020gaussian}.
For the optimizer, Adam, we use a weight decay of $1\text{e-}6$ and an exponential decay learning rate scheduler (with gamma $= 0.98$, and a minimum learning rate of $1\text{e-}6$). We clip the L2 gradient norm of all our models at $1$, which is important due to the unnormalized targets. 
We use LayerNorm \cite{ba2016layer} for the MLP, while all other models do not use any network normalization -- these configurations were found empirically to be superior for the respective models.

\subsection{Eigenvector centrality analysis}
\label{sec:appendix_lgcn}
\input{figures/latex/globals_eigv_centrality}
Following \cite{cachay2021world}, we analyze the learned adjacency matrix of L-GCN (see Fig. \ref{fig:learned_adj} for the explicit structure), via the node eigenvector centrality score method.
See \cite{cachay2021world} for details of the method.
A high centrality score for a node translates to the node being important within the graph. In the particular case of a GCN the score reflects into the core message-passing forward-pass \cite{kipf2017gcn}, since the node propagates its information to a greater extent than other nodes.
Our eigenvector centrality analysis shows that L-GCN learns to assign a high importance to the global node, see Fig. \ref{fig:globals_eigv}. The figure shows how the score for the global node converges across differently seeded runs to a very high score of over 0.8 (in the later epochs no other node has a score that surpasses 0.5, and most nodes have scores lower than 0.05).
This underlines the importance of using the non-spatial \emph{globals} information that contains important boundary conditions like the sun angle as well as surface type and geographical related information.

%% file: appendix/tables/invars.tex
\begin{table}
\centering
\caption{Definition of all the physical \embf{input variables} (var.), and whether they are part of the \emph{globals} (G), \emph{layers} (Lay), or \emph{levels} (Lev) input type. The storing scheme for the input variables is described in \ref{sec:storing_scheme}.
A cross in the Clear-sky column indicates that the corresponding variable is only used for experiments with clear-sky conditions.}
\medskip
\label{tab:input_vars}
\begin{tabular}{|c|c|c|c|c|c|} %\begin{tabular}{@{} *5l @{}}
\toprule
Var. Name & Definition & G & Lay & Lev & Clear-sky \\
\midrule
\midrule
$shtj$ & Eta coordinate at layer interfaces (levels) & & & \cmark & \\
$trow$ & Temperature at levels & & & \cmark &  \\
$shj$ & Eta coordinate at layer mid-point & & \cmark & &  \\
$dShj$ & Layer thickness in eta coordinate & & \cmark & & \\
$dz$	& Geometric thickness of the layer & & \cmark & & \\
\hline
$height$	& Geometric height of a level & & & \cmark & \\
$tlayer$ &	Temperature at layer mid-point & & \cmark & & \\
$temp\_diff$ &	Temperature difference between levels & & \cmark & & \\
$qc$ &	Water vapour  & & \cmark & &  \\
$ozphs$ &	Ozone  & & \cmark & &  \\
\hline
$co2rox$ &	CO2 concentration & & \cmark & &  \\
$ch4rox$ &	CH4 concentration & & \cmark & &  \\
$n2orox$&	N2O concentration & & \cmark & &  \\
$f11rox$&	CFC11 concentration & & \cmark & &  \\
$f12rox$&	CFC12 concentration & & \cmark & &  \\
\hline
$rhc$ & Relative humidity & & \cmark & & \cmark \\
$aerin$ & Aerosol mass mixing ratios &  & \cmark & &\cmark \\
$sw\_ext\_sa$	& Solar extinction coefficient for stratospheric aerosols &  & \cmark & & \cmark\\
$sw\_ssa\_sa$	& Solar single scattering albedo for stratospheric aerosols &  & \cmark & &\cmark \\
$sw\_g\_sa$ & Solar asymmetry for stratospheric aerosols &  & \cmark & &\cmark  \\
$lw\_abs\_sa$	& Thermal absorptivity for stratospheric aerosols &  & \cmark & &\cmark  \\
\hline
$pressg$ &	Surface pressure & \cmark & & &  \\
$level\_pressure$ & Level pressure & & & \cmark &  \\
$layer\_pressure$ & Layer pressure & & \cmark & &  \\
$layer\_thickness$ & Layer thickness in pressure & & \cmark & &  \\
$gtrow$   &   Grid-mean surface temperature & \cmark & & &  \\
\hline
$oztop$   &   Ozone above the top of the model & \cmark & & &   \\
$cszrow$ &	Cosine of the solar zenith angle & \cmark & & &  \\
$emisrow$ & 	Grid-mean surface emissivity & \cmark & & &  \\
$salbrol$ & 	Grid-mean all-sky surface albedo & \cmark & & &  \\
$csalrol$ & 	Grid-mean clear-sky surface albedo & \cmark & & &  \\
\hline
$emisrot$ & 	Surface emissivity for each surface tile & \cmark & & & \\
$gtrot$ & 	Surface temperature for each surface tile& \cmark & & & \\
$farerot$ & 	Fraction of grid of each surface tile & \cmark & & &   \\
$salbrot$ & 	All-sky surface albedo for each surface tile & \cmark & & &    \\
$csalrot$ & 	Clear-sky surface albedo for each surface tile & \cmark & & & \\
\hline
$x$-$cord$ & see \ref{sec:appendix_geo_info} & \cmark & & & \\
$y$-$cord$ & 	see \ref{sec:appendix_geo_info} & \cmark & & & \\
$z$-$cord$ & 	see \ref{sec:appendix_geo_info} & \cmark & & & \\
\bottomrule
\end{tabular}
\end{table}

%% file: appendix/tables/outvars.tex
\begin{table}
\centering
\caption{Definition of all the physical \embf{output variables} (var.).
The naming is the same for both pristine- and clear-sky, but are stored in different subdirectiories: \emph{outputs\_pristine/} and \emph{outputs\_clear\_sky/} respectively.
The profile type column indicates whether the variable profile is across the levels or layers of the column.
}
\medskip
\label{tab:output_vars}
\begin{tabular}{@{} *5l @{}}
\toprule
Var. Name & Definition & Profile type &  Unit \\
\midrule
\midrule
$rsuc$ & Up-welling shortwave (solar) flux  & levels & $W/m^2$\\
$rsdc$ & Down-welling shortwave (solar) flux &  levels & $W/m^2$\\
$hrlc$	& Solar heating rate profile & layers & $K/s$ \\
\hline
$rluc$ & Up-welling longwave (thermal) flux & levels & $W/m^2$ \\
$rldc$ & Down-welling longwave (thermal) flux & levels & $W/m^2$\\
$hrlc$	& Thermal heating rate profile & layers & $K/s$ \\
\bottomrule
\end{tabular}
\end{table}

%% file: figures/latex/models_per_year_long.tex
\begin{figure}[htb]
\centering
 \begin{subfigure}[b]{.48\linewidth}
    \centering
    \includegraphics[width=.99\textwidth]{figures/figs/yearly/model_SW_FLUX_RMSE_per_year.png}
    \caption{Vertical avg. RMSE}\label{fig:avg_rmse}
    \end{subfigure}%
  \begin{subfigure}[b]{.48\linewidth}
    \centering
    \includegraphics[width=.99\textwidth]{figures/figs/yearly/model_SW_FLUX_MBE_per_year.png}
    \caption{Vertical avg. MBE}\label{fig:avg_mbe}
 \end{subfigure} \\
  \begin{subfigure}[b]{.48\linewidth}
    \centering
    \includegraphics[width=.99\textwidth]{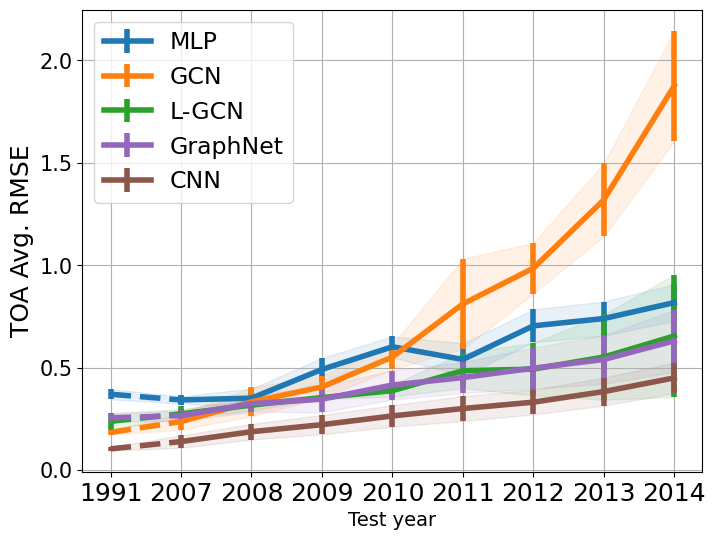}
    \caption{TOA RMSE}\label{fig:toa_rmse}
  \end{subfigure}%
  \begin{subfigure}[b]{.48\linewidth}
    \centering
    \includegraphics[width=.99\textwidth]{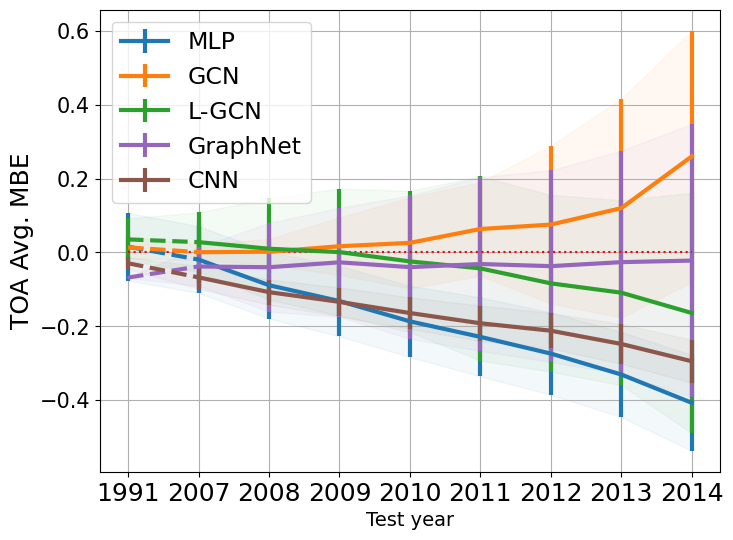}
    \caption{TOA MBE}\label{fig:toa_mbe}
 \end{subfigure} \\
  \begin{subfigure}[b]{.48\linewidth}
    \centering
    \includegraphics[width=.99\textwidth]{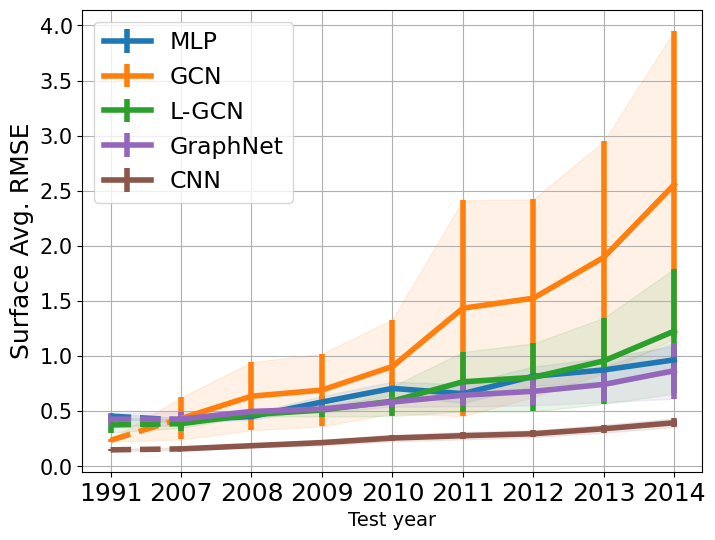}
    \caption{Surface RMSE}\label{fig:surface_rmse}
  \end{subfigure}%
  \begin{subfigure}[b]{.48\linewidth}
    \centering
    \includegraphics[width=.99\textwidth]{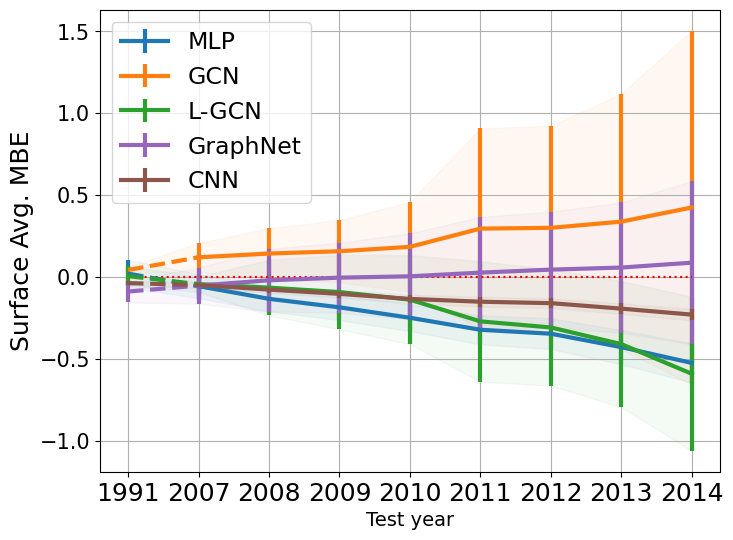}
    \caption{Surface MBE}\label{fig:surface_mbe}
 \end{subfigure} 
  \caption{Performance as a function of the test year at different levels for our baseline models. (Fig. \ref{fig:avg_rmse}) and (Fig. \ref{fig:avg_mbe}) show the errors vertically averaged over all levels of a column (profile). 
 The TOA errors are shown in (Fig. \ref{fig:toa_rmse}) \& (Fig. \ref{fig:toa_mbe}) and the error at the surface is presented in (Fig. \ref{fig:ood_rmse}) \& (Fig. \ref{fig:ood_mbe}). Apart from the superior performance of CNN, it's interesting to note the miniscule mean bias error (MBE) of the GraphNet. which is an important property for climate simulations.}
  \label{fig:yearly_long}
\end{figure}

%% file: figures/latex/model_per_ood_long.tex
\begin{figure}[htb]
\centering
 \begin{subfigure}[b]{.48\linewidth}
    \centering
    \includegraphics[width=.99\textwidth]{figures/figs/ood/bar_plots/model_FLUX_RMSE_OOD_bars.png}
    \caption{Vertical avg. RMSE}\label{fig:ood_avg_rmse}
    \end{subfigure}%
  \begin{subfigure}[b]{.48\linewidth}
    \centering
\includegraphics[width=.99\textwidth]{figures/figs/ood/bar_plots/model_FLUX_MBE_OOD_bars.png}
    \caption{Vertical avg. MBE}\label{fig:ood_avg_mbe}
 \end{subfigure} \\
  \begin{subfigure}[b]{.48\linewidth}
    \centering
    \includegraphics[width=.99\textwidth]{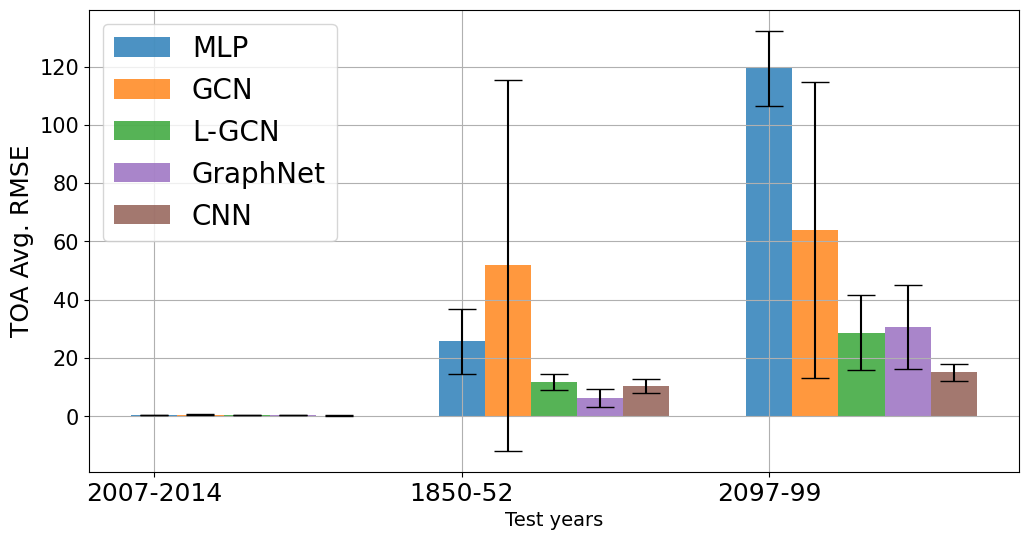}
    \caption{TOA RMSE}\label{fig:ood_toa_rmse}
  \end{subfigure}%
  \begin{subfigure}[b]{.48\linewidth}
    \centering
    \includegraphics[width=.99\textwidth]{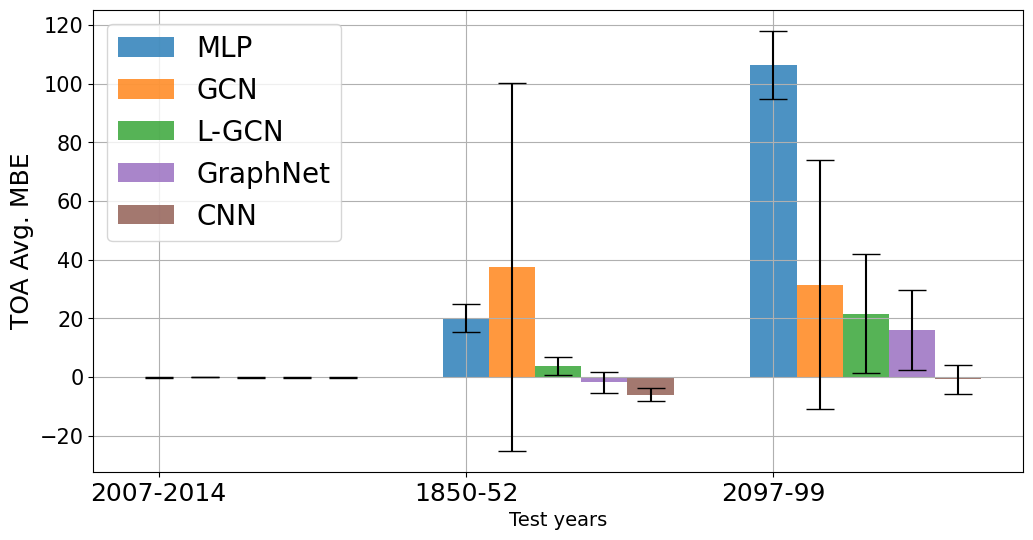}
    \caption{TOA MBE}\label{fig:ood_toa_mbe}
 \end{subfigure} \\
  \begin{subfigure}[b]{.48\linewidth}
    \centering
    \includegraphics[width=.99\textwidth]{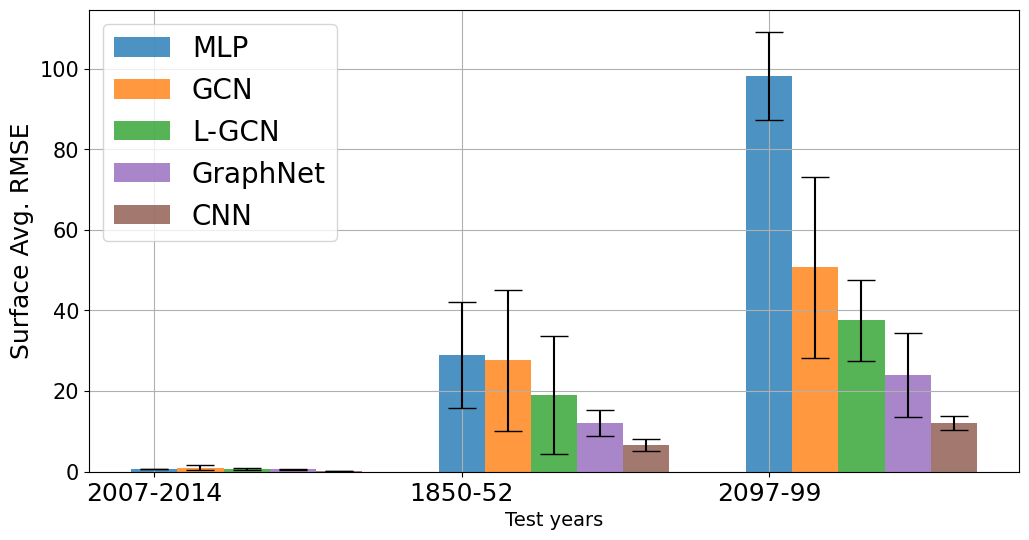}
    \caption{Surface RMSE}\label{fig:ood_surface_rmse}
  \end{subfigure}%
  \begin{subfigure}[b]{.48\linewidth}
    \centering
    \includegraphics[width=.99\textwidth]{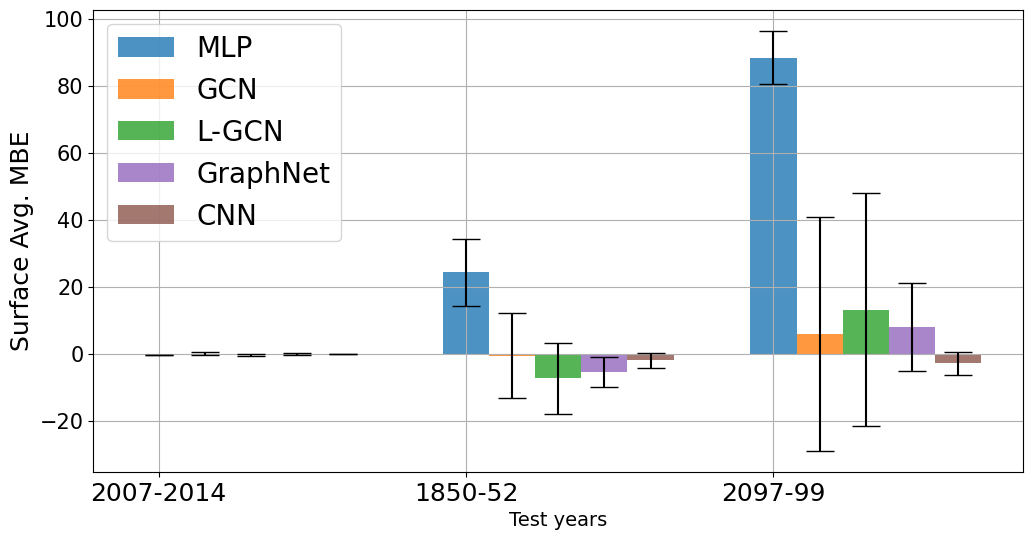}
    \caption{Surface MBE}\label{fig:ood_surface_mbe}
 \end{subfigure} 
 
 \caption{Performance as a function of the test year at different levels for our baseline models. (Fig. \ref{fig:ood_avg_rmse}) and (Fig. \ref{fig:ood_avg_mbe}) show the errors vertically averaged over all levels of a column (profile). 
 Apart from the vertically averaged errors, it's important to calculate the errors in top of the atmosphere (TOA) and surface levels as they're used for the calculation of heating rates from the predicted radiative flux.
 The TOA errors are shown in (Fig. \ref{fig:ood_toa_rmse}) \& (Fig. \ref{fig:ood_toa_mbe}) and the error at the surface is presented in (Fig. \ref{fig:ood_surface_rmse}) \& (Fig. \ref{fig:ood_surface_mbe}). As expected, CNNs and Graph-based models (L-GCN \& GraphNet) are far more superior in all the levels compared to the MLPs for whom the error in future predictions is higher by and order of a magnitude.
          }
  \label{fig:ood_long}
\end{figure}

%% file: figures/latex/globals_eigv_centrality.tex
\begin{figure}
     \centering
     \includegraphics[width=.95\linewidth]{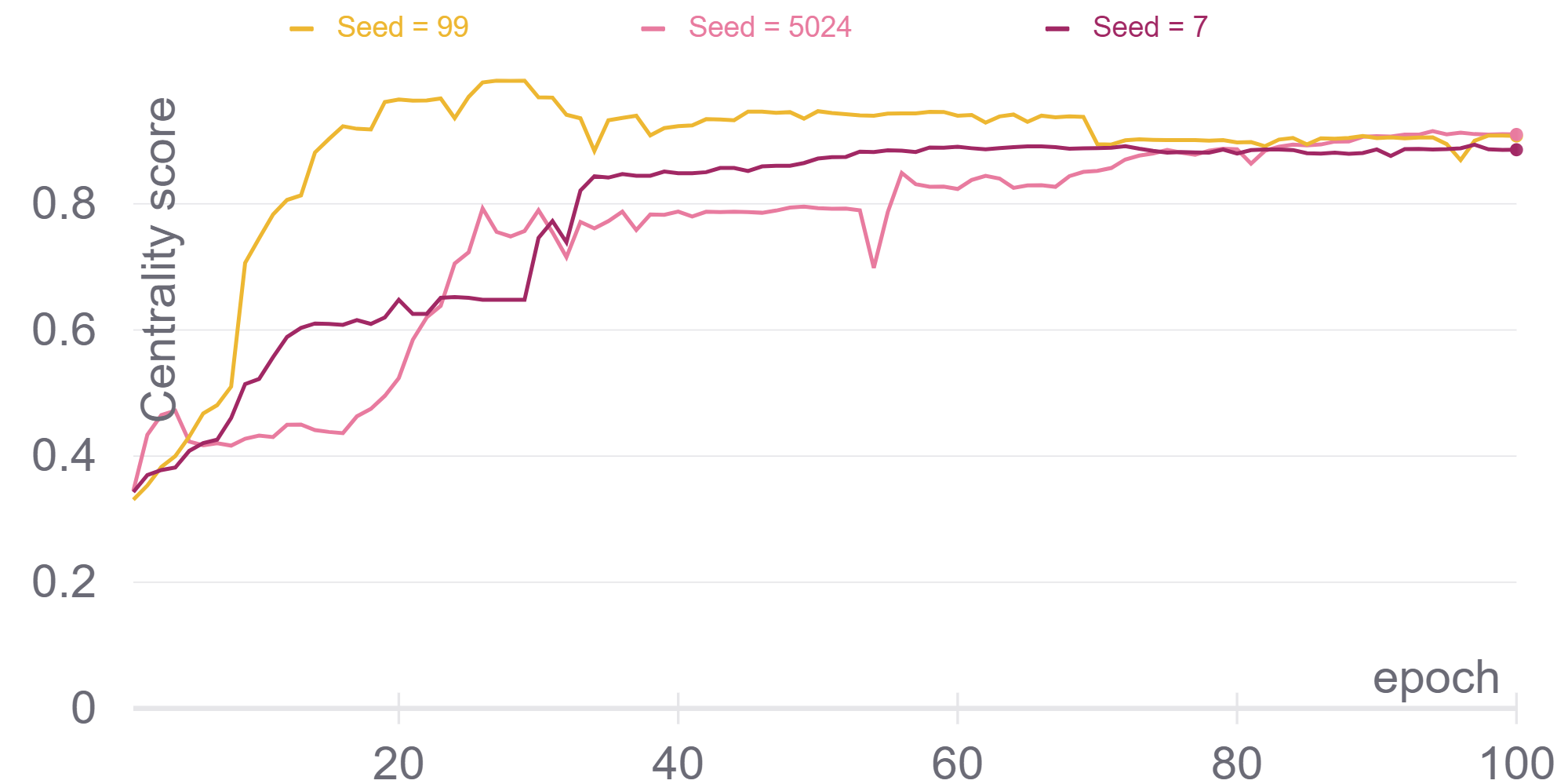}
    \caption{L-GCN, a graph convolutional network with learnable adjacency matrix, 
    learns to give high importance to the global node, which contains boundary conditions information, as measured by its high eigenvector centrality score (for the learned adjacency matrix).
    We plot this score as a function of the epoch for all three differently seeded runs of L-GCN.
    See Appendix \ref{sec:appendix_lgcn} for more discussion.
    }
  \label{fig:globals_eigv}
\end{figure}